%% file: colm2025_conference.tex
\documentclass{article} 
\usepackage[preprint]{colm2025_conference}

\usepackage{microtype}
\usepackage{graphicx}
\usepackage{subfigure}
\usepackage{hyperref}
\usepackage{url}
\usepackage{booktabs}
\usepackage{adjustbox}
\usepackage{amssymb}
\usepackage{algorithm}
\usepackage{algpseudocode}

\usepackage{lineno}

\definecolor{darkblue}{rgb}{0, 0, 0.5}
\hypersetup{colorlinks=true, citecolor=darkblue, linkcolor=darkblue, urlcolor=darkblue}

\title{Read Quietly, Think Aloud: Decoupling Comprehension and Reasoning in LLMs}

\author{Yuanxin Wang, \& Ganesh Venkatesh \\
AppliedML, Cerebras\\
\texttt{\{michael.wang, ganesh.venkatesh\}@cerebras.net} \\
}

\begin{document}

\ifcolmsubmission
\linenumbers
\fi

\maketitle

\newcommand{\ignore}[1]{}
\newcommand{\fixme}[1]{#1}
\definecolor{DarkGreen}{RGB}{1,50,32}
\newcommand{\hl}[1]{\textbf{\textcolor{DarkGreen}{#1}}}
\newcommand{\readq}{\scalebox{.9}[1.0]{\textsc{ReadQ}}}
\newcommand{\readqe}{\scalebox{.9}[1.0]{\textsc{ReadQEmb}}}
\newcommand{\readqb}{\scalebox{.9}[1.0]{\textsc{ReadQBuddy}}}
\newcommand{\readqba}{\scalebox{.9}[1.0]{\textsc{ReadQBuddyAdapt}}}
\newcommand{\algcomment}[1]{\scalebox{.8}[1.0]{\textit{\texttt{#1}}}}

\begin{abstract}
Large Language Models (LLMs) have demonstrated remarkable proficiency in understanding text and generating high-quality responses. However, a critical distinction from human cognition is their typical lack of a distinct internal `reading' or deliberation phase before `speaking' (i.e., generating text). Humans often engage in silent reading to comprehend context and formulate thoughts prior to articulation. This paper investigates methods to imbue LLMs with a similar capacity for internal processing.

We introduce and evaluate techniques that encourage LLMs to `read silently.' Our findings indicate that even a straightforward approach, such as providing the model with an initial contextual prompt or `reading space' before it begins predicting subsequent tokens for the final output, can yield significant performance improvements. We further enhance this concept by developing a `reading buddy' architecture, where an auxiliary component silently processes the input and provides refined contextual insights to the primary generation model. These approaches aim to foster deeper understanding from LLMs so that they can produce better reasoned responses, moving them one step closer to more human-like text processing. Our results indicate that these simple techniques can provide surprisingly strong impact on accuracy with multiple point accuracy boost. 
\end{abstract}

\input{intro}
\input{motivation}
\input{approach}

\input{results}

\input{conclusion}

\bibliography{colm2025_conference}
\bibliographystyle{colm2025_conference}

\appendix
\input{appendix}

\end{document}

%% file: intro.tex
\section{Introduction}
\label{sec:intro}

\ignore{
1. LLMs ability to answer complex questions have grown remarkably in recent times through enhanced training recipes - in particular Online RL (such as GRPO) as well as test-time compute.
2. The big breakthrough here is giving the model more space to think by having a "thinking" phase where it lays out the reason before answering teh question.
3. However, there is currently no way to encourage to internalize teh context and the question before answering it. Previous works such as Read Again~\cite{readagain} likely point to the situation where model spending more time reading and internalizing before starting to produce response can lead to better responses.
4. We provide a novel training recipe and associated model architecture changes to encourage teh model to ``read quietly'' before it starts to produce thinking traces and the final response.
5. We believe this is a promising area of research and potential approach to help model produce better reasoned approaches without costing extra think tokens at runtime.
}

The ability of Large Language Models (LLMs) to tackle complex questions has grown remarkably in recent times~\cite{o1,deepseekr1,qwq,gemini}. This progress has been driven by enhanced training recipes, such as online reinforcement learning techniques like Generative Reward Policy Optimization~\cite{grpo}, as well as by allocating more computation at inference time. A key breakthrough common to many of these advancements is giving the model dedicated space to ``think'' — a phase where it explicitly lays out a chain of reasoning before providing the final answer. This has proven to be a powerful mechanism for improving the fidelity and logic of generated responses.

However, the current paradigm, with its focus on the explicit ``thinking'' trace, largely overlooks the crucial preceding step: the model's initial comprehension and internalization of the context and query. There are currently no widespread methods to explicitly encourage an LLM to spend more time processing the input before it begins to speak. Prior work, such as Read Again~\cite{readagain}, indicates that providing models with more time to "read" and internalize information before they are required to produce a response can lead to significant quality improvements, highlighting a clear opportunity for exploration.

In this work, we address this gap by proposing a novel training recipe and corresponding model architecture changes. These modifications are specifically designed to encourage the model to ``read quietly'' before it generates its thinking traces and final answer. We believe that formally incorporating a silent reading phase is a promising research direction. By providing the model with a superior initial comprehension of the context, we posit that this ``silent reading'' phase can amplify the benefits of test-time ``thinking'' and could be a promising future direction of exploration.

%% file: motivation.tex
\section{Motivation and Related Work}
\label{sec:motivation}

\ignore{
1. Humans produce reasoned responses by "understanding the content" and then thinking through the response.
2. Much of the current focus on reasoning in LLMs targets the second aspect -- "thinking through" the response with CoT, inference-time techniques and more recently reasoning models.
3. Very little work on encouraging the LLM to "quietly" read the context to internalize the context as prep for thinking through and producing the response.
4. Our work proposes two simple techniques to encourage the model to think quietly before speaking. We believe the encouraging results make a strong case for reconsidering LLM training recipe and model architecture to encourage "reading quietly" and "thinking/responding" phases.
}
\begin{figure}[b]
\vspace{-10pt}
\centering
\includegraphics[width=\columnwidth]{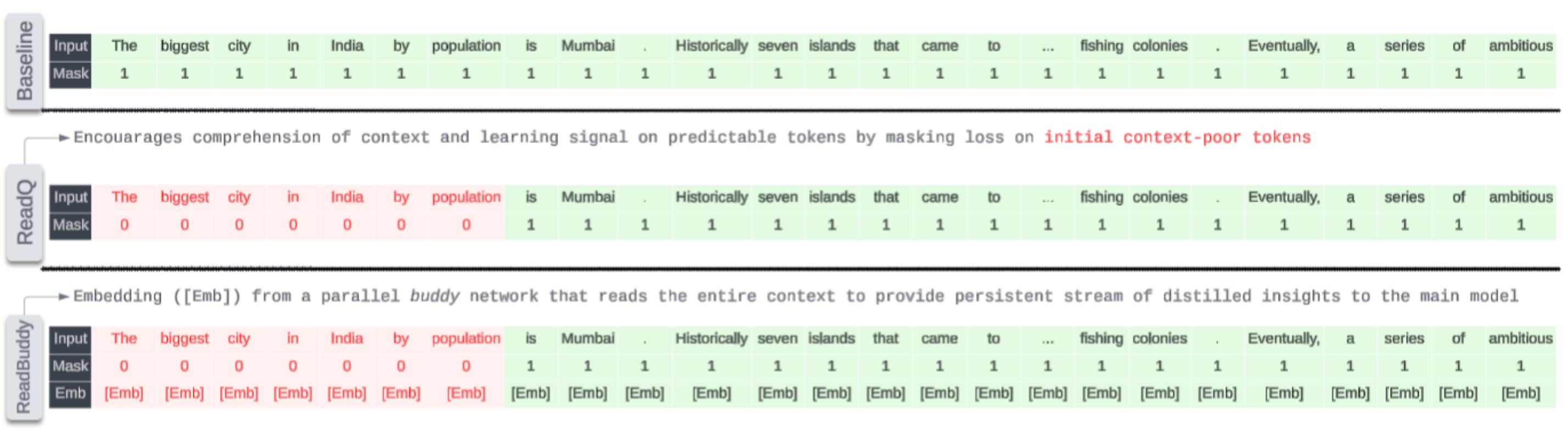}
\caption{\label{fig:approach} \textbf{An illustration of our proposed \readq\ and \readqb\ techniques}. }
\vspace{-5pt}
\end{figure}
We hypothesize that an effective process for generating reasoned responses comprises two distinct stages. The first is a `silent reading' phase, dedicated to internalizing provided information to build a comprehensive understanding of the context and the query's demands. This is followed by a second, `response formulation' phase, where the agent consciously thinks through the problem to construct a coherent answer. We posit that explicitly modeling these two stages is a critical, yet overlooked, aspect of improving machine reasoning.

Much of the recent focus on improving reasoning in Large Language Models (LLMs) has predominantly targeted the second stage of this process—the explicit ``thinking'' or articulation of reasoning. Techniques such as Chain-of-Thought~\cite{cot} prompting, test-time compute~\cite{cepo, mcts}, and specialized reasoning models~\cite{deepseekr1,qwq} are all designed to improve how an LLM structures and generates its thought process as part of the output. However, comparatively little attention has been given to the foundational first stage: encouraging the model to ``read quietly'' to internalize and synthesize the context before initiating the reasoning and generation phase. This represents a significant gap, as the quality of the final response is fundamentally dependent on the depth of the initial understanding.

Our work proposes first steps towards addressing this challenge. We propose two simple yet effective techniques designed to encourage the model to perform this silent reading step (Section~\ref{sec:approach}). The encouraging results from our approach make a compelling case for reconsidering the standard LLM training and architectural paradigms. We believe our findings highlight a valuable opportunity to explicitly incorporate distinct ``reading quietly'' and ``thinking/responding'' phases into future model designs, potentially leading to more robust and human-like efficient reasoning capabilities.

\ignore{
1. Another line of recent work is understanding the importance of each token in a sample and weighting them accordingly.
2. For a given sample, it is very unlikely LLM would be able to predict the first few tokens correctly given it has no context. For example, consider the statement "The biggest city in India is XYZ". LLM model has no easy way to predict the --> biggest transition or "biggest --> city", but it likely can predict is -> XYZ given it has the context. 
3. Esp as the LLM gets strong and has strong predictive power over much of the sample (esp in multi-epoch settings), the only tokens with high loss would be the initial tokens where the model has little to no context. Given that, it seems futile in those settings to update weights based on those initial token losses.
4. Our work naturally addresses this concern as well in a simple and elegant way.
}

A related line of recent work has begun to explore the differential importance of tokens within a training sample~\cite{rho}, often with the goal of weighting their contribution to the loss function accordingly. A fundamental challenge in standard autoregressive training: the initial tokens of a sequence are inherently difficult, if not impossible, to predict correctly given the lack of preceding context. For example, in a sequence like "The largest city in US by population is New York," the model has very little signal to predict "largest" from "The," but has a very strong signal to predict "New York" once the context "...US by population is" has been processed.

As models become more powerful and are trained for multiple epochs, they achieve very low loss on the predictable, context-rich portions of a sample. Consequently, the training signal becomes disproportionately dominated by the high loss from the unpredictable initial tokens. Arguably, forcing the model to update its weights based on these high-variance, context-poor predictions is an inefficient, if not futile, application of the learning objective. Our "silent reading" framework naturally addresses this concern in a simple and elegant manner. By designating an initial period for comprehension where the next-token prediction loss is not applied, our work effectively instructs the model to focus on understanding before predicting, thereby sidestepping the problem of penalizing the model for failing at an inherently speculative task.

%% file: approach.tex
\section{Approach}
\label{sec:approach}

\ignore{
1. We propose two simple techniques - `read quietly' and `read buddy'
2. Read Quietly encourages the model to spend the time during the first few tokens reading silently before starting to produce the response. We do by by loss masking the initial tokens. By doing so, at the minimum we are reducing the impact of training on high-variance, context-poor tokens that are hard to predict and in the ideal case, giving the model space to develop ability to read quietly. 
3. We recognize that the above approach only provides this benefit for the first few tokens but the later tokens do not get the benefit of reading quietly to build comprehension. To that end, we extend the model with a Read Buddy that provides the model with comprehension from reading the context.
}

To encourage LLMs to better understand context before generating a response, we introduce two simple yet effective techniques shown in Figure~\ref{fig:approach}: Read Quietly (\readq) and Read Buddy (\readqb). The first technique modifies the training process to create a ``silent reading'' window at the beginning of a sequence, while the second introduces an architectural enhancement to provide persistent comprehension throughout the generation process.

\subsection{\readq: Loss Masking for Initial Comprehension}
\label{sec:readq}

Our first technique, \readq, encourages the model to dedicate the initial moments of processing to silent reading rather than immediate prediction. We implement this by applying loss masking to the first K tokens of the input sequence during training. Specifically, for a given sequence, the standard autoregressive next-token prediction loss is not calculated for the first K predictions after each bos token.

This approach provides a twofold benefit. At a minimum, it addresses the issue of training on high-variance, context-poor initial tokens. These early tokens are inherently difficult to predict, and the gradients derived from them can be noisy and inefficient for learning. By masking this portion of the loss, we prevent the model from being penalized for failing at a speculative task. In the ideal case, we hypothesize that this pressure-free window gives the model the opportunity to develop an ability to ``read quietly'' - that is, to use these initial steps to build a more robust internal representation of the context before commencing the generation task.

\begin{table*}[]
\caption{\readq\ and \readqb\ deliver significant gains on multiple benchmarks. SW16/SW64 - sliding window of 16/64 for \readqb}
\label{tab:results}
\vskip -0.5in
\begin{center}
\begin{small}
\begin{sc}
\begin{adjustbox}{width=\columnwidth}
\begin{tabular}{lccccc}
\toprule
Experiment & Arc Challenge & Hellaswag & OpenbookQA & PubmedQA & Winogrande \\
\midrule
Llama 3.1 3B IT & 37.2 & 57.56 & 34.0 & 65.0 & 62.35 \\
\readq & 45.82 & 72.64 & 40.8 & 75.4 & 69.22 \\
\readqb & 49.06 & 73.03 & 40.8 & 75.6 & 68.82 \\
\hline
& & \textbf{Ablations} & & &\\\midrule
Llama 3.2 3B Base & 37.97 & 61.98 & 35.4 & 67.6 & 64.4 \\
+\readq & 45.56 & 71.6 & 40.0 & 70.8 & 67.4 \\\midrule
Llama 3.1 3B IT & 37.2 & 57.56 & 34.0 & 65.0 & 62.35 \\
+\readqb\ + SW16 & 47.18 & 72.64 & 40.8 & 75.4 & 70.4 \\
+\readqb\ + SW64 & 47.7 & 73.0 & 41.0 & 75.6 & 70.32 \\
\bottomrule
\end{tabular}
\end{adjustbox}
\end{sc}
\end{small}
\end{center}
\vskip -0.1in
\end{table*}
\subsection{\readqb: Extending Comprehension with an Auxiliary Module}
\label{sec:readqb}

We recognize that the \readq\ technique primarily benefits the initial phase of generation. However, for long and complex inputs, later tokens in the generated response could also benefit from a continuously available understanding of the preceding context. The comprehension built during the first few steps may dilute as the generation sequence grows.

To that end, we propose \readqb, a model extension designed to provide persistent comprehension. This approach involves an auxiliary ``buddy'' module that reads the entire input context in parallel, processes the information and provide a semantic representation of the context to the primary generation model at each step. This ensures that the core insights from the ``silent reading'' phase are not lost and can inform the entire reasoning and response-generation process, from the first token to the last. In our particular implementation (Algorithm~\ref{alg:approach}), \readqb\ is a LLM model and the semantic representation is the embedding from the second last layer. The input to the main model is the sum of the semantic representation from the buddy model and output of embedding layer from the main model.

%% file: results.tex
\section{Results}
\label{sec:results}

\ignore{
\begin{enumerate}
    \item Contd. pretrain with a ``read'' mode
    \item ``Read Buddy'' further amplifies benefits
    \item Base vs SFT model -- applicable to both
    \item Sliding window for read buddy
    \item At different model scales
\end{enumerate}
1. We present analysis of our proposal on Llama 3.2 3B Instruct model with Llama 3.2 1B as read buddy where applicable. We also analyze the ReadQ approach at a much larger scale of Llama 3.1 70B and find interesting patterns.
2. Table~\ref{tab:results} presents our main results and shows a consistent win for readq and readbuddy. These are very early results and especially for readbuddy, it is likely our recipe is under-training them and not giving model sufficient time to fully leverage teh extra embedding information. But it is encouraging to see the benefits even in this simple setting.
3. To better understand this, we present a couple of ablations: i) use base model instead of the instruct model, ii) using sliding window in the readbuddy to further enrich each representation. We notice that in each case there are benchmarks that get a significant multi-point benefit pointing to the fact that a better training recipe could realize further gains.
4. Finally, we show the accuracy on Llama 3.1 70B (Table~\ref{tab:resmodelsize} and here again see similar story where the wins are consistent across multiple benchmarks and very significant in some cases (for example 8 percentage point jump for MedQA).
}

We now present the empirical validation for our proposed \readq\ and \readqb\ methods, with full details on our datasets and training recipe provided in Appendix~\ref{sec:appendix}. For each comparison, the baseline is model trained on same dataset without our techniques. Our findings demonstrate consistent performance improvements in multiple settings and confirm that these benefits hold when scaling to larger models and domain-specific dataset. 

\subsection{Main Results for \readq\ and \readqb}
\label{sec:resmain}

Our primary experiments evaluate our proposed methods, \readq\ and \readqb, on the Llama 3.2 3B Instruct model~\cite{llama}. For the \readqb\ configuration, a Llama 3.2 1B model serves as the ``buddy'' that provides contextual summaries. Table~\ref{tab:results} presents our main findings across a suite of benchmarks. We observe a consistent performance improvement for both the \readq\ and \readqb\ configurations compared to the baseline model. It is worth noting that these are early results. We suspect our current training recipe may under-train the \readqb\ variant in particular, not allowing the main model sufficient time to fully leverage the information from the buddy's embeddings. The gains seen even in this simple setting are therefore highly encouraging and point to significant potential.

\subsection{Ablation Studies}
\label{sec:ablation}

To better understand the behavior of our methods, we conduct two key ablation studies. First, we apply our methods to the base model instead of the instruct-tuned version to isolate the effect of instruction tuning. Second, we experiment with enriching the \readqb's representations by incorporating a local sliding window. When extending \readqb\ with sliding window, we notice that while overall averages may be similar, specific benchmarks see significant, multi-point benefits. This suggests that a more tailored training recipe for different model types and configurations could unlock substantial further gains.

\subsection{Scalability to Larger Models and Specialized Datasets}
\label{sec:res70b}

Finally, to demonstrate that the benefits of our approach hold when scaling to a larger model trained on a specialized dataset, we evaluate \readq\ on a much larger Llama 3.1 70B model trained on a data mixture targeting scientific domain. The results, summarized in Table~\ref{tab:resmodelsize}, tell a similar story. The performance gains are consistent across multiple benchmarks, confirming the scalability of our method. Notably, we see a very significant impact on some tasks, such as an 8 percentage point jump in accuracy on MedQA~\cite{medqa}, highlighting the potential of our approach in complex reasoning scenarios.

\begin{table*}[]
\caption{Accuracy boost for larger Llama 3.1 70B model on scientific tasks}
\label{tab:resmodelsize}
\vskip -0.05in
\begin{center}
\begin{small}
\begin{sc}
\begin{tabular}{lccccc}
\toprule
Experiment & MedQA & MedMCQA & PubmedQA & GPQA & MMLU \\
\midrule
Llama 3.1 70B & 66.8 & 69.1 & 79.6 & 37.37 & 78.9 \\
+ ReadQ & 74.7 & 69.7 & 80.0 & 38.38 & 79.2 \\
\bottomrule
\end{tabular}
\end{sc}
\end{small}
\end{center}
\vskip -0.1in
\end{table*}

%% file: conclusion.tex
\section{Conclusion}
\label{sec:conclusion}

\ignore{
1. We explore a new direction -- encouraging models to read quietly and comprehend before starting the reasoning.
2. Early results promising with some significant accuracy boosts with a relatively simple training recipe -- this makes a strong case for exploring this phase to augment and boost the reasoning power of current day areasoning models.
}

This paper introduced `reading quietly' a simple yet powerful technique to ensure LLMs comprehend context before generating a response, leading to superior reasoning. The natural next step is to combine this foundational `reading' phase with the advanced `thinking' capabilities of state-of-the-art models. This synthesis of silent comprehension and explicit reasoning offers a promising candidate for the future of artificial intelligence.

%% file: appendix.tex
\section{Appendix}
\label{sec:appendix}

\subsection{Training Details for Llama 3.2 3B and Llama 3.1 70B}

\ignore{
1. Trained on a combination of Fineweb~\cite{fineweb}, Deepmind~\cite{deepmind}, Pile~\cite{pile}, Cosmopedia~\cite{cosmo} and UltraBooks~\cite{utb}. We train for 4096 steps using a batch size of 1000.
2. Our focus was on our techniques and so we did not optimize for the dataset or training recipe. In particular, given the low loss of the model during continued pretraining (around 1.5 or lower), hypothesize that the model was already likely good at these datasets and does not benefit much in terms of new information. We believe a better curated dataset and ablated training recipe could possibly realize larger gains.
3. For ReadBuddy, we train in two phases where the first phase is only 500 steps and only trains projection of buddy model's embedding to same dimension as main model. Rest of the training is same.
}

We trained for 4096 steps using a global batch size of 1000. The training corpus was a diverse mixture of publicly available datasets, including Fineweb~\cite{fineweb}, selected data from Deepmind Math~\cite{deepmind}, Pile~\cite{pile}, Cosmopedia~\cite{cosmopedia}, and UltraBooks~\cite{utb}.

It is important to note that our experimental focus was on the efficacy of our proposed techniques, and as such, we did not perform an exhaustive optimization of the dataset mixture or the overall training recipe. We conducted continued pretraining on a highly capable base model, which maintained a low training loss (around 1.5 or lower) throughout this process. We hypothesize that the model had likely already assimilated much of the information in these datasets during its original training, potentially limiting the knowledge-based gains. We believe that future work with a more carefully curated dataset and a fully ablated training recipe could realize even larger performance improvements from our methods.

For the ReadBuddy model, we employed a two-phase training process. The first phase consisted of 500 steps where we exclusively trained the projection layer responsible for mapping the buddy model's output embedding to the same dimension as the main model's hidden state. This initial, brief phase allows for a stable alignment of the two components. After these 500 steps, the rest of the model parameters were unfrozen, and the training proceeded for the remaining duration under the same configuration as our other experiments.

For Llama 3.1 70B, we trained on a combination of pretraining and supervised fine-tuning dataset focused primarily on scientific domain. We train for 1361 steps using batch size of 960. This dataset mixture and training recipe was developed in context of another exploration and we layer our innovations (\readq\ and \readqb) on it without any change to validate the general applicability of our technique.

\subsection{\readqb\ Algorithm}

\begin{algorithm}
\caption{\label{alg:approach} \readqb\ Algorithm Pseudo-code}
\begin{algorithmic}[1]
\State \textbf{INPUT:} T$_{in}$, T$_{out}$, M$_{gen}$, M$_{read}$, C
\State \textbf{OUTPUT:} $\mathcal{L}$
\State \algcomment{\#T$_{in}$, T$_{out}$: Text input and output}
\State \algcomment{\#M$_{read}$, M$_{gen}$, E$_{gen}$ C: Read, Gen, Embedding and Conn. Model}
\State T$_{pred}$ = M$_{gen}$(E$_{gen}$ (T$_{in}$) + C(M$_{read}$(T$_{in}$)) 
\State $\mathcal{L}$ = $\mathcal{CE}$(\texttt{lm\_head}(T$_{pred}$), T$_{out}$) \algcomment{\#Next token loss} 
\end{algorithmic}
\end{algorithm}